\def\paperID{0000}
\def\confName{CVPR}
\def\confYear{2024}

 \def\authorBlock{
Lin Liu,~~Yutong Wang,~~Jiahao Chen,~~Jianfang Li,\\~~Tangli Xue,~~Longlong Li,~~Jianqiang Ren,~~Liefeng Bo \\ 

   Tongyi Lab,~~Alibaba Group\\
   
    {\tt\small \{lorrain.ll, yutong.yutongwang, peter.cjh,  wuhui.ljf, xuetangli.xtl, } \\
 {\tt\small  longer.lll, jianqiang.rjq, liefeng.bo\}@alibaba-inc.com}
}

\newif\ifreview 
\newif\ifarxiv \newcommand{\arxiv}{\arxivtrue}
\newif\ifcamera 
\newif\ifrebuttal 

\arxiv  

\pdfoutput=1
\documentclass[10pt,onecolumn,letterpaper]{article}
\ifreview \usepackage[review]{cvpr} \fi
\ifarxiv \usepackage[pagenumbers]{cvpr} \fi
\ifrebuttal \usepackage[rebuttal]{cvpr} \fi
\ifcamera \usepackage{cvpr} \fi

\usepackage{graphicx}	
\usepackage{amsmath}	
\usepackage{amssymb}	
\usepackage{booktabs}
\usepackage{times}
\usepackage{microtype}
\usepackage{epsfig}
\usepackage[table,xcdraw,dvipsnames]{xcolor}
\usepackage{caption}
\usepackage{float}
\usepackage{placeins}
\usepackage{color, colortbl}
\usepackage{stfloats}
\usepackage{enumitem}
\usepackage{tabularx}
\usepackage{xstring}
\usepackage{multirow}
\usepackage{xspace}
\usepackage{url}
\usepackage{subcaption}
\usepackage{xcolor}
\usepackage[hang,flushmargin]{footmisc}

\ifcamera \usepackage[accsupp]{axessibility} \fi

% ...

\ifarxiv  \fi

\newcommand{\R}[1]{{%
    \textbf{%
        \ifstrequal{#1}{1}{\textcolor{red}{R#1}}{%
        \ifstrequal{#1}{2}{\textcolor{blue}{R#1}}{%
        \ifstrequal{#1}{3}{\textcolor{magenta}{R#1}}{%
        \ifstrequal{#1}{4}{\textcolor{teal}{R#1}}{%
                           \textcolor{cyan}{R#1}%
        }}}}%
    }%
}}

\usepackage{xr-hyper}

\makeatletter
\newcommand*{\addFileDependency}[1]{
  \typeout{(#1)}
  \@addtofilelist{#1}
  \IfFileExists{#1}{}{\typeout{No file #1.}}
}

\makeatother
\newcommand*{\myexternaldocument}[1]{
    \externaldocument{#1}
    \addFileDependency{#1.tex}
    \addFileDependency{#1.aux}
}

\definecolor{cvprblue}{rgb}{0.21,0.49,0.74}
\usepackage[pagebackref,breaklinks,colorlinks,citecolor=cvprblue]{hyperref}
\usepackage[capitalize]{cleveref}
\crefname{section}{Sec.}{Secs.}
\crefname{table}{Table}{Tables}
\crefname{figure}{Fig.}{Figs.}

\ifarxiv \crefname{appendix}{App.}{Apps.}
\else \crefname{appendix}{Suppl.}{Suppls.} \fi

\usepackage{indentfirst}
\unless\ifarxiv \myexternaldocument{_supplementary} \fi

\begin{document}
\title{Make-A-Character 2: \\ Animatable 3D Character Generation From a Single Image}
 
\author{\authorBlock}
\maketitle

\begin{abstract}


This report introduces Make-A-Character 2, an advanced system for generating high-quality 3D characters from single portrait photographs, ideal for game development and digital human applications. Make-A-Character 2 builds upon its predecessor by incorporating several significant improvements for image-based head generation. We utilize the IC-Light method to correct non-ideal illumination in input photos and apply neural network-based color correction to harmonize skin tones between the photos and game engine renders. We also employ the Hierarchical Representation Network to capture high-frequency facial structures and conduct adaptive skeleton calibration for accurate and expressive facial animations. The entire image-to-3D-character generation process takes less than 2 minutes. Furthermore, we leverage transformer architecture to generate co-speech facial and gesture actions, enabling real-time conversation with the generated character. These technologies have been integrated into our conversational AI avatar products. 

\end{abstract}
\section{Introduction}
\label{sec:intro}

Make-A-Character\cite{ren2023makeacharacter} introduced a system that generates 3D characters based on text descriptions. Although text is effective and lightweight, it struggles to describe the precise nuances of facial features, limiting its controllability on detailed attributes. As the saying goes, a picture is worth a thousand words, images can convey far more detailed information than text, allowing users to generate characters with greater intuitiveness and control.

Given the aforementioned content, we introduce Make-A-Character 2, a system for 3D characters generation from single images. We wish to generate a 3D character with consistent appearance (\emph{e.g.}, face, hairstyle, skin color) given a frontal face portrait photo, and the 3D character is equipped with sophisticated underlying skeleton and ready for animation. The Make-A-Character 2 inherits key properties from Make-A-Character, such as highly-realistic rendering, full-body completed and industry-compatible. Further more, we propose the following major improvements:

\noindent {\bf Portraits Illumination Harmonization.}
To ensure the high quality of the generated 3D head, ideal input portraits should be captured under optimal lighting conditions. This includes uniform lighting that avoids asymmetry, excessive shadows, or overexposure. However, daily portraits often fail to meet these strict requirements. To address this issue, we utilize the latest diffusion-based illumination editing method IC-Light\cite{iclight} to correct the lighting in the given photos, bringing them to an ideal illumination state for subsequent operations.

\noindent {\bf Color Correction in Game Engine.}
We aim to ensure that the generated 3D character maintains consistent lighting and skin tone with the input photo. However, predicting the lighting environment from an input portrait and replicating it in render engine is challenging. The final color appearance in a game engine cannot be precisely predicted from a diffuse texture map alone due to various influencing factors, such as lighting setup, material properties, reflections, subsurface scattering, and ambient occlusion. We draw inspiration from \cite{shi2019face} and address the challenge of non-differentiability in game engines by training a neural network to learn the color discrepancies between diffuse maps and their corresponding renders from the game engine. After the network successfully models this color offset, we utilize the learned model to adjust the diffuse map, allowing the render engine to produce the desired color outcome more accurately.

\noindent {\bf Detailed Face Reconstruction.}
Our initial head geometry, reconstructed guided by facial dense landmarks, often lacks high-frequency details. To address this limitation, we leverage HRN\cite{lei2023hierarchical} facial geometry to augment the surface details of our initial model. For essential facial regions, such as around the eyes and lips, we increase the density of landmark predictions, enabling to capture intricate nuances with greater precision.


\noindent {\bf Facial Skeleton Calibration.}
According to industry standards, the creation of 3D facial assets necessitates a well-designed skeletal structure and specialized rigging to achieve satisfactory animation outcomes. In our work, we conducted skeleton fitting on the reconstructed facial meshes, utilizing several neutral face models sourced from MetaHuman, which had already undergone comprehensive rigging processes as described in \cite{metahumandnacalibration}. This approach guarantees that each facial mesh produced through the aforementioned geometric reconstruction algorithm is equipped with a customized skeletal configuration.

\noindent {\bf Real-Time Speech-driven Animation.} 
Although there are many researches~\cite{liu2024emageunifiedholisticcospeech, zhu2023taming,ijcai2023p650,chen2024diffsheg,zhang2024semtalkholisticcospeechmotion} addressing the problem of audio-driven motion generation, the  produced expression and gesture animations are far from satisfactory. They mainly adopt ARKit blendshapes to represent the facial expressions. The ARKit blendshapes are not able to express detailed mouth shapes. As for the gesture, their generated results are not restricted and the motions are jittering. In order to produce realistic motions, we follow Metahuman's control rig to animate facial expressions and employ two senior animators to revise motion capture data. Instead of generating gestures freely, we predict indices of motion pieces, which are intentionally designed by animators. This allows our system to generate natural and realistic body motions.

\section{Related Work}
\label{sec:related}
\noindent {\bf Single-View 3D Reconstruction.} Reconstructing a 3D object from a single image is an ill-posed problem that typically relies on specific prior knowledge. Inspired by the Scaling Law, which demonstrates notable success in natural language processing (NLP) tasks through the utilization of scalable network architectures and large training datasets, recent works \cite{hong2023lrm, tochilkin2024triposr} propose a large-scale 3D reconstruction model based on the Transformer architecture. These models aim to extract a generalized 3D prior from extensive datasets, thereby enabling the prediction of Neural Radiance Field (NeRF) representations from a single image. While these methods demonstrate remarkable reconstruction quality and generalization capabilities, they face significant challenges in the realm of 3D avatar generation, which demands sophisticated structures to ensure animatability and compatibility with existing industry CG pipelines.

\noindent {\bf Single-View 3D Head Generation.} The 3D Morphable Model (3DMM) \cite{blanz1999morphable} captures the variability inherent in human face by representing facial geometries as a linear combination of blendshapes. This approach has become a foundational technique in 3D face reconstruction. By fitting the model parameters to 2D images, it enables face reconstruction from single images.

With the advent of deep learning, researchers \cite{deng2019accurate, tewari2017mofa} have developed methods to leverage neural networks for predicting 3D face shapes directly from images. To better capture fine details, many approaches aim to improve the 3D Morphable Model (3DMM)-based framework by introducing nonlinear 3DMM \cite{tran2018nonlinear}, deformation maps generation \cite{lei2023hierarchical}, and displacement maps generation \cite{feng2021learning} for higher-quality geometry. Regarding texture, the traditional linear combination of texture bases often results in a blurred texture map. Thus, techniques like differentiable rendering 
 \cite{lei2023hierarchical} or directly unwrapping the input image into UV space \cite{lin2021meingame} are more effective alternatives.

Advancements in neural 3D representations have led to the development of a variety of models for detailed human head reconstruction. The Neural Parametric Head Models (NPHM) \cite{giebenhain2023learning} effectively use a signed distance field (SDF) to achieve high-fidelity capture of complete human head geometry. Despite their detailed geometrical focus, NPHMs do not support appearance modeling or hair reconstruction. In contrast, Rodin \cite{wang2023rodin} employs diffusion models to create 3D digital avatars using neural radiance fields (NeRF), which incorporate both geometry and appearance. Additionally, the 3D Gaussian Parametric Head Model (GPHM) \cite{xu2024gphm} leverages 3D Gaussian to deliver photorealistic rendering quality and real-time performance. Nonetheless, several challenges persist, including integrating the Gaussian head model with a full body and adapting these models to varying lighting conditions.

\noindent {\bf Speech-driven 3D Gesture Generation.} Gesture generation is a complex task that requires understanding speech, gestures, and their relationships. To address this, BEAT ~\cite{10.1007/978-3-031-20071-7_36} presents a high-quality, multi-modal motion data set captured from 30 speakers. Along with the data set a Cascaded Motion Network is proposed to generate gestures from speech. Based on this data set, EMAGE ~\cite{Liu_2024_CVPR} and ProbTalk ~\cite{Liu_2024_CVPRProbtalk} utilize masked transformer to predict gestures from audio in the latent space of VQ-VAE. Although improved results are observed, the motion diversity is restricted. 
Diffusion based models ~\cite{ijcai2023p650, chen2024diffsheg, zhu2023taming} are also explored in the literature of speech-driven motion generation. Compared to the previous methods, these  could generate diverse motions. However, the  results produced by all of these methods are far from industrial standards. The predicted motions are usually jittering. Instead of unrestricted generation, we directly predict the indices of predefined motions, which are designed by senior animators. Based on this, it is able to generate realistic gestures motions.

\noindent {\bf Speech-driven 3D Face Animation.} Speech-driven 3D face animation focuses on creating realistic facial animations from the input speech signals. This field has developed for a long history and achieved significant progress, which can be broadly categorized into visme-based and deep learning-based methods. Visme-based methods primarily focused on establishing mapping relationships from phonemes to visemes \cite{taylor2012dynamic,edwards2016jali,bao2023learning}. These methods allow for explicit control and can be easily integrated into industrial animation pipelines. However, they have some disadvantages: they mainly concentrate on lip region animations and lack comprehensive strategies for animating the entire face. Additionally, the expressiveness of these methods is limited, making it difficult to capture detailed lip movements.
Recently, deep learning-based methods have been increasingly researched. Karras et al. \cite{karras2017audio} proposed an deep neural network to learn a mapping from input waveforms to 3D vertex coordinates of a face model. VOCA \cite{cudeiro2019capture} utilizes an encoder-decoder network where the encoder learns to transform audio features to a low-dimensional embedding and the decoder maps this embedding into a high-dimensional space of 3D vertex displacements. Faceformer \cite{fan2022faceformer} also leverages an encoder-decoder model, but with Transformer-based architecture and autoregressively generates a sequence of animated 3D face meshes from input raw audio. To avoid over-smoothed facial motions, Codetalker \cite{xing2023codetalker} adopts a pre-trained VQ-VAE motion prior. MeshTalk \cite{richard2021meshtalk} achieves highly realistic motion synthesis results for the entire face based a categorical latent space that disentangles audio-correlated and audio-uncorrelated information. SelfTalk \cite{peng2023selftalk} involve a self-supervision framework to exchange cross-modals information to generating realistic and accurate lip movements with lipreading comprehensibility.
However, the methods mentioned above are all based vertex offset, usually lacking components like teeth, tongue, and eyelashes, leading to flaws during animation. Additionally, they are difficult to integrate into existing animation production pipelines. However, these approaches can serve as valuable references.

\section{3D Character Generation}
\label{sec:character_genneration}

\begin{figure*}[tb]
\centering
\includegraphics[width=0.98\textwidth]{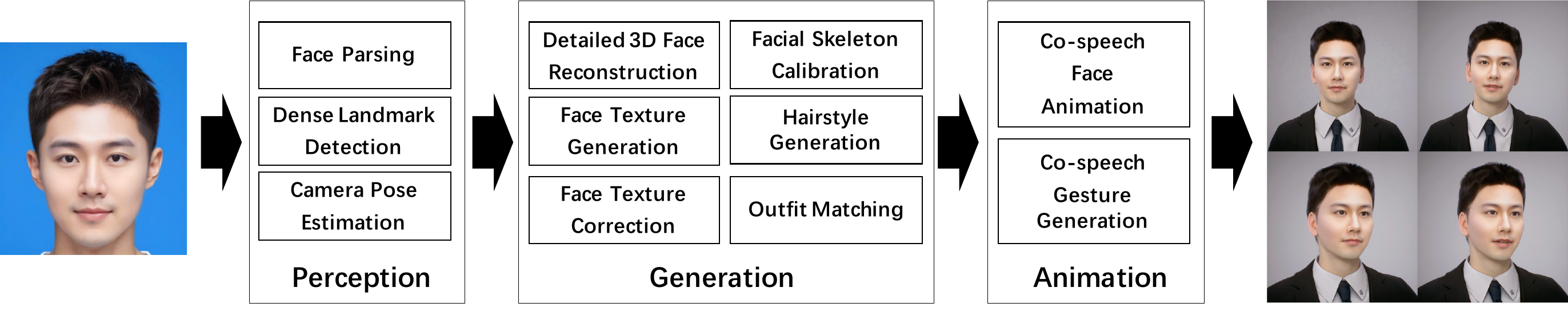}
\caption{Pipeline for generating animatable 3D characters from a single image.}
\label{fig:generation_pipeline}
\vspace{-15pt}
\end{figure*}

\subsection{Geometry Generation}
\subsubsection{Hierarchical Facial Details Transfer}

\begin{figure}[tb]
    \centering
    \includegraphics[width=0.7\linewidth]{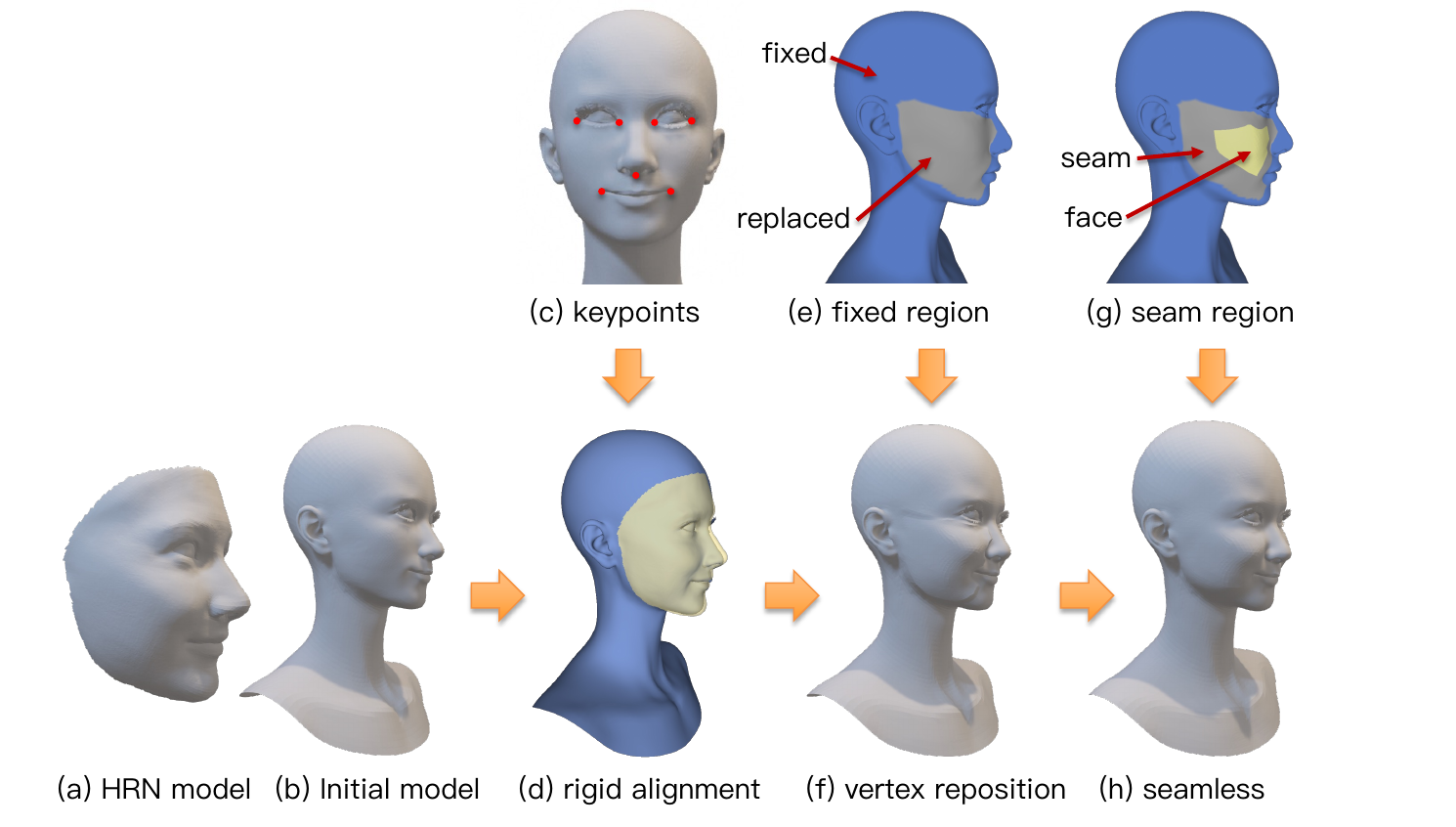}
    \caption{The facial details transfer pipeline. We align the HRN geometric model (a) with the initial geometric model (b) through a rigid alignment process (d) based on seven key points indicated in (c). Following this, we fix the positions of the points within the blue region, as depicted in (e) of our initial model, and subsequently transfer the HRN facial details to the replaceable gray region also shown in (e), leading to the creation of a replaced head model with HRN face details (f). Finally, we determine the gray transition region as shown in (g) and apply a smoothing technique to attain a final model (h) that seamlessly integrates facial details and exhibits a natural appearance.}
    \label{fig:face_detail_transfer}
\end{figure}

Unlike our prior work\cite{ren2023makeacharacter} which relied on textural descriptions, this research uses a single portrait image as input, requiring more accurate and detailed face generation. Our prior triplane-based method, which used total variation loss for smoothness, struggled to reconstruct intricate facial details like wrinkles and dimples in the initial head geometry. Lei et al.~\cite{lei2023hierarchical} proposed a Hierarchical Representation Network (HRN) enabling highly accurate single-image facial reconstruction. We utilize this HRN to transfer detailed facial features to our initial head model, overcoming previous geometric limitations. To align the detail-rich HRN geometry with our model, we apply a rigid transformation based on 7 facial landmarks, followed by details geometry transfer. However, direct transfer can induce artifacts at the junctions between the detail-enhanced facial areas and the fixed head region. To ensure a seamless transition and achieve a natural appearance, we employ a smoothing operation at these junctions. The complete pipeline is shown in Figure \ref{fig:face_detail_transfer}.

\begin{figure}[tb]
    \centering
    \includegraphics[width=0.6\linewidth]{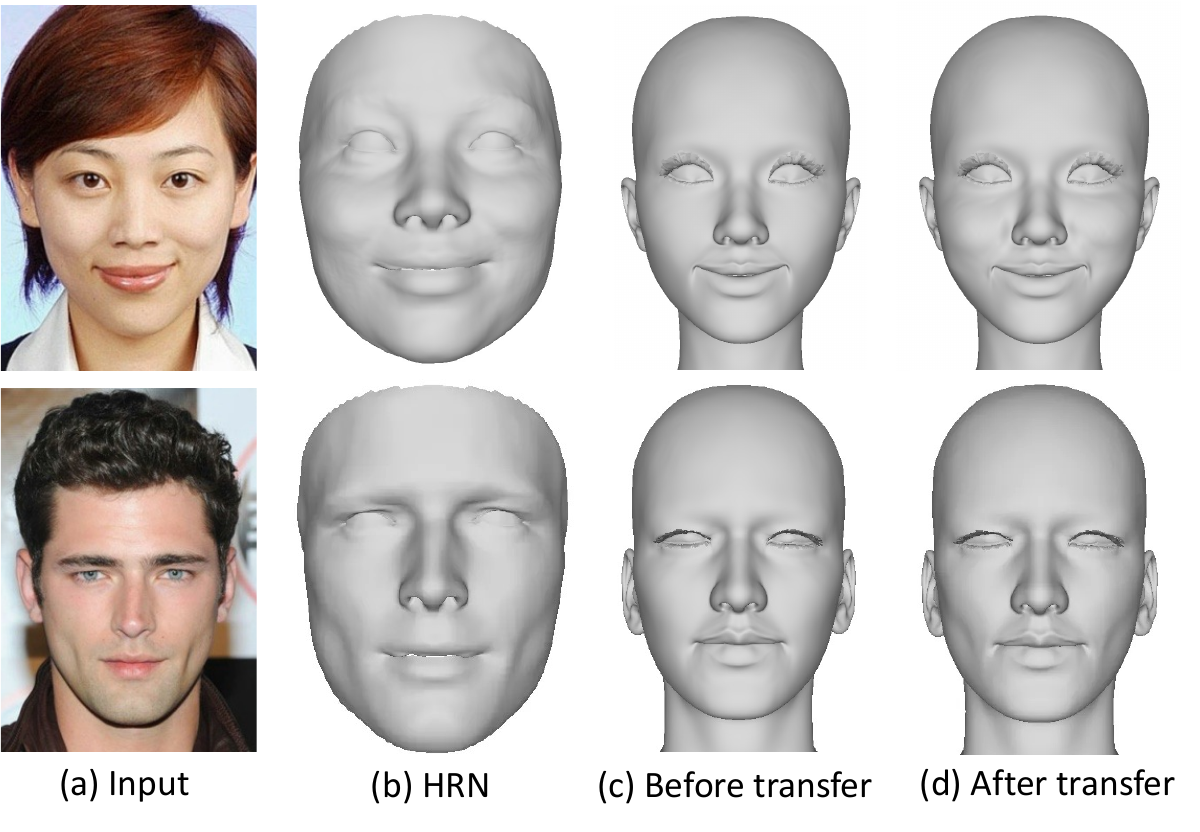}
    \caption{Results with facial details generated using our facial detail transfer pipeline.}
    \vspace{-10pt}
    \label{fig:face_detail_transfer_result}
\end{figure}

Figure \ref{fig:face_detail_transfer_result} showcases the effectiveness of our HRN-guided facial detail transfer. Using images from the SCUT-FBP5500\cite{liang2017SCUT} as input, the figure highlights a substantial enhancement in facial details. Notably, the pre-transfer faces appear smooth and lack fine features, whereas the post-transfer faces exhibit detailed characteristics, such as nasolabial folds and dimples.

\begin{figure}[tb]
    \centering
    \includegraphics[width=0.6\linewidth]{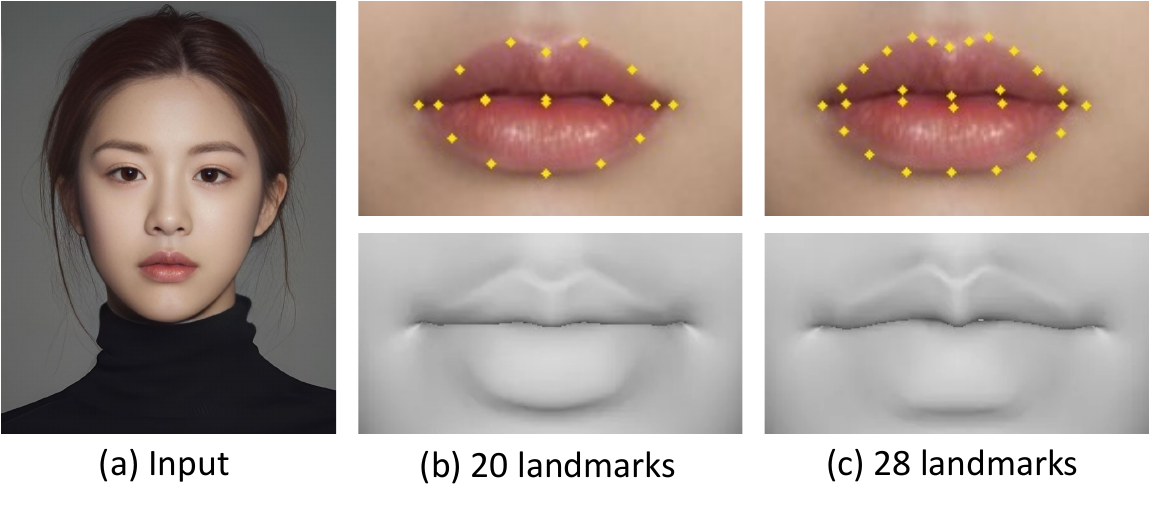}
    \caption{An example of mouth geometric results landmarks obtained by expanding the mouth landmarks}
    \label{fig:mouth_geo}
\end{figure}

\begin{figure}[tb]
    \centering
    \includegraphics[width=0.6\linewidth]{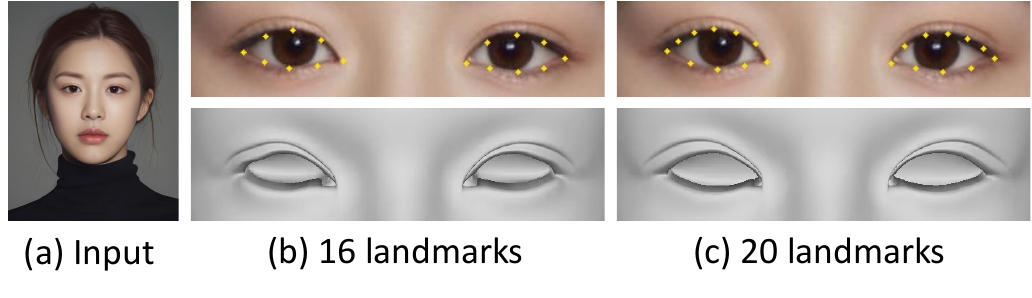}
    \caption{An example of eyes geometric results landmarks obtained by expanding the eyes landmarks}
    \label{fig:eyes_geo}
\end{figure}

The complexity of the eyes and mouth requires a more detailed representation than the original landmarks could provide. Therefore, we increased the landmark count to 20 for the eyes and 28 for the mouth. This crucial modification significantly improves the geometric precision of these facial features, resulting in superior alignment with the input image data. Figures \ref{fig:mouth_geo} and \ref{fig:eyes_geo} visually demonstrate this improvement.


\begin{figure}
    \centering
    \includegraphics[width=0.6\linewidth]{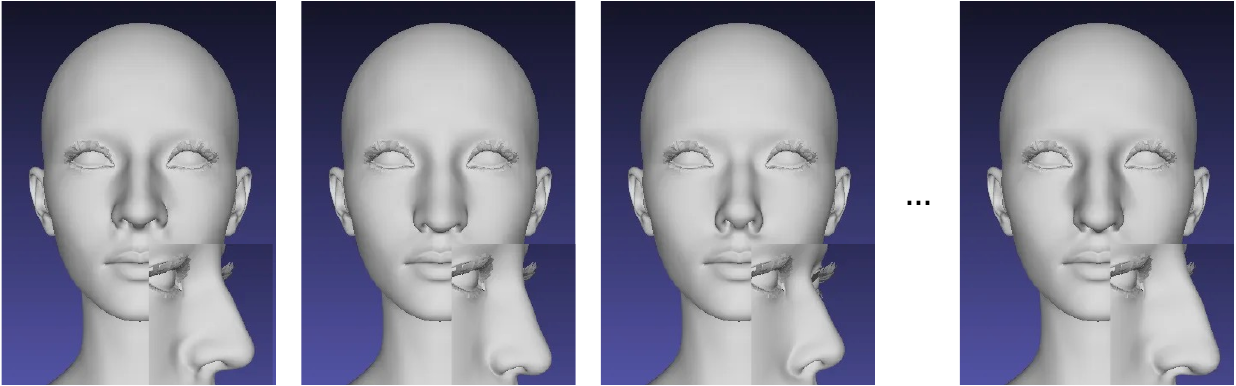}
    \caption{Nose Shape Basis Set Derived from the Basel Face Model.}
    \label{fig:nose_geo}
    \vspace{-10pt}
\end{figure}

Because the nose and adjacent facial areas have similar skin tones, accurately detecting nose landmarks for shape reconstruction is a difficult task. To improve the precision of nose shape reconstruction, we leverage the shape knowledge encoded within the Basel Face Model (BFM). Specifically, we augment the nose representation with 80 shape bases from the BFM to obtain the final geometry $V$:


\begin{equation}
V = triplane\left [v, u, :\right ] + \sum_{i=0}^{80} S_i\alpha _i
\end{equation}
As shown in Figure \ref{fig:nose_geo}, the nose shape of $S_i$ is consistent with BFM shape basis, but the topology is geometrically identical to the triplane representation,  and $\alpha_i$ is the nose shape coefficient.

\subsubsection{Facial Skeletons Calibration}

Once high-fidelity human face meshes are generated, accurate skeleton(\emph{i.e.}, joint, bone) binding is crucial for ensuring the realistic and nuanced animation driving performance. 
We achieve this by fitting the skeleton (joints and bones) from a pre-rigged, neutral-shaped face mesh onto the newly generated target model.

 Our approach begins by implementing the forward propagation of a skeleton-driven skinning system. This system uses a hierarchical, multi-branch, loop-free tree structure to organize hundreds of skeleton nodes, mirroring the rig logic systems of MetaHuman\cite{metahumanriglogic}. Each node stores its local translation and rotation relative to its parent. Once a skeleton moves, we update the absolute positions of all descendant nodes of this skeleton using forward kinematics. These displacements in absolute coordinate, combined with Linear Blend Skinning (LBS) weights, determine the displacements of mesh vertices and deform the mesh. Since our geometry reconstruction maintains consistent face-vertex topology, we can compute the vertex differences between the target mesh and the deformed mesh on a per-vertex basis. These differences are then used as the loss function for skeleton fitting:

\begin{equation}
L_v = \sum_{n=1}^{N} \left\| \mathbf{v}_i^{\text{current}} - \mathbf{v}_i^{\text{target}} \right\|_2^2
\end{equation}
where $N$ denotes the number of vertices in the mesh, $\mathbf{v}_i^{\text{current}}$ and $\mathbf{v}_i^{\text{target}}$ are the $(x, y, z)$ coordinates of the i-th vertex in the current deformed mesh and the target mesh, respectively. We aim to move the current vertices towards their corresponding target positions.

\begin{figure}[tb]
    \centering
     \includegraphics[width=0.3\linewidth]{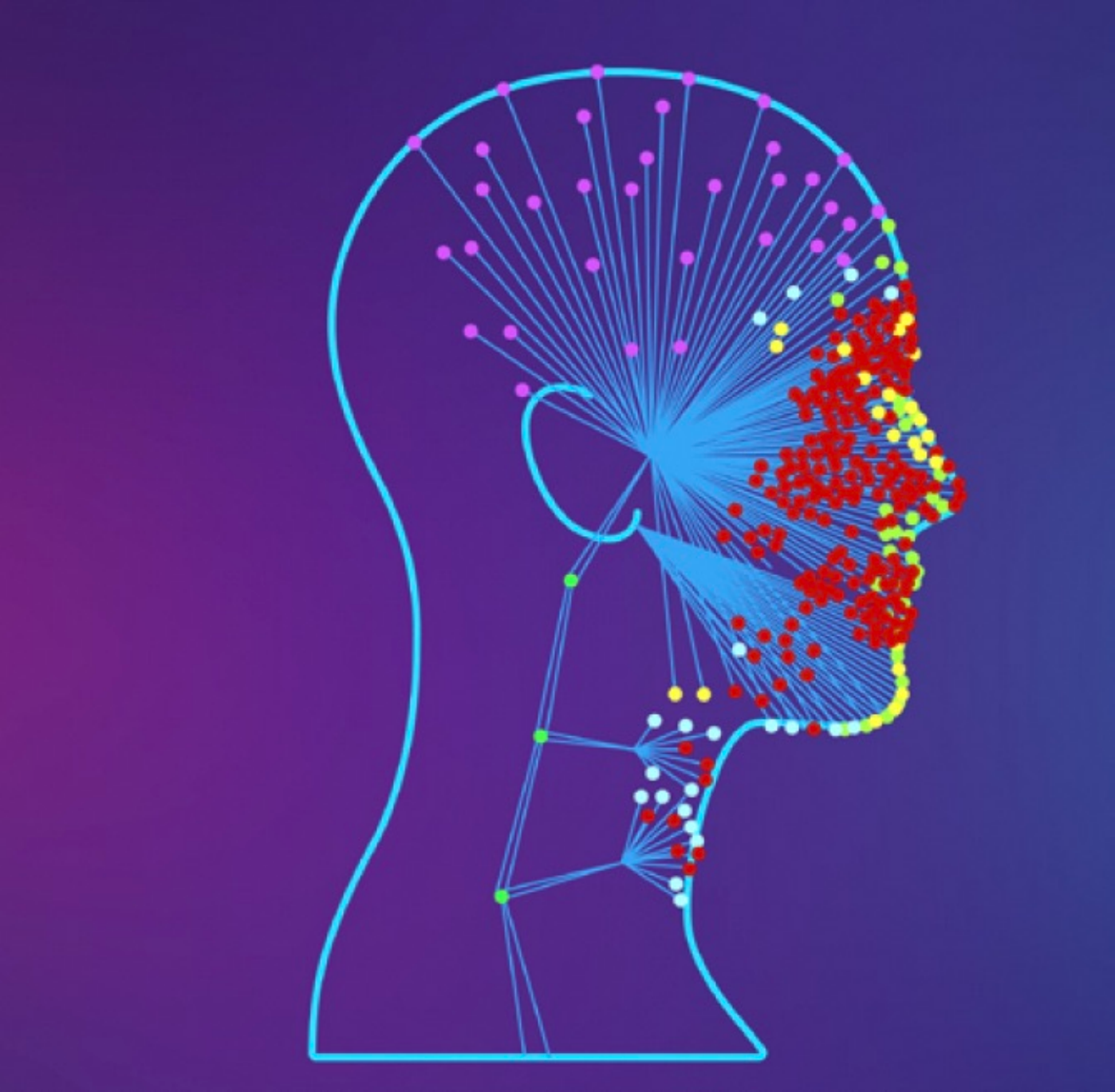}
    \includegraphics[width=0.7\linewidth]{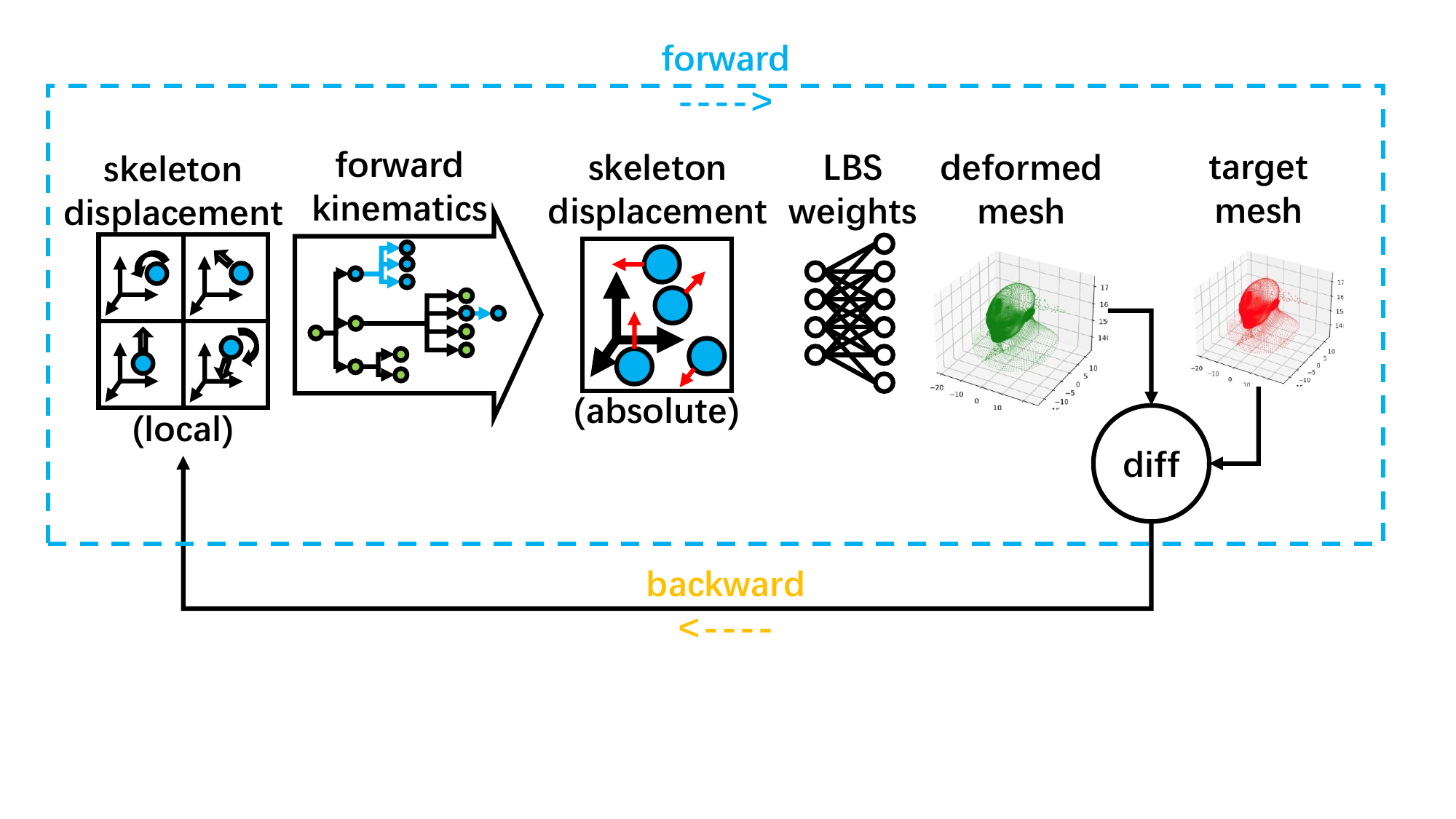}
   \vspace{-50pt}
    \caption{The skeleton calibration pipeline.}
    \label{fig:skeleton_calibration_pipeline}
    \vspace{-10pt}
\end{figure}

 Leveraging the complete forward path from skeleton displacement to mesh vertex displacement, we can directly utilize automatic differentiation to optimize skeleton parameters. This process iteratively adjusts the skeleton position, aiming to minimize the loss and drive the current mesh closer to the target shape. This iterative fitting pipeline, visualized in Figure \ref{fig:skeleton_calibration_pipeline}, continues until convergence is achieved.

We also observe significant spatial overlap between the leaf node skeletons and the mesh vertices on both the neutral and rigged face meshes. We denote the index set of these overlapping leaf skeletons as $\Phi$:
\begin{equation}
M(i)=j,  i \in  \Phi
\end{equation}
\begin{equation}
\mathbf{s}_i=\mathbf{v}_{M(i)} , i \in  \Phi
\end{equation}

the function $M(.)$ maps each index of a skeleton to the corresponding index of its associated vertex in the mesh. Therefore, the positional differences between these overlapping skeleton nodes and their mesh counterparts are used as a supervision component in the fitting process:


\begin{equation}
L_s = \sum_{i \in \Phi}  \left\| \mathbf{s}^{\text{current}}_i - \mathbf{v}^{\text{current}}_{M(i)} \right\|_2^2
\end{equation}

the total loss function is given by $L_{total}$= $L_v$ + $L_s$.

Through practical experimentation, we observed that enabling updates for all skeleton parameters during the optimization process can lead to unreasonable skeleton configurations. Specifically, the optimizer tends to induce exaggerated movements in the skeleton in an attempt to align the deformed mesh with the target mesh. To mitigate this issue, we impose a constraint by allowing updates only for the leaf node skeletons within the skeleton tree system.

Simply fine-tuning the skeletons alone is insufficient to perfectly match the target shape (the loss cannot be reduced to zero). Therefore, our process first applies the residual offset as a blend shape to the neutral face mesh. Subsequently, skeletons are adjusted according to the results of the above solution. This two-step calibration enables a well-rigged face mesh.


\begin{figure}[tb]
    \centering
    \begin{subfigure}{0.2\linewidth}
        \includegraphics[width=\linewidth]{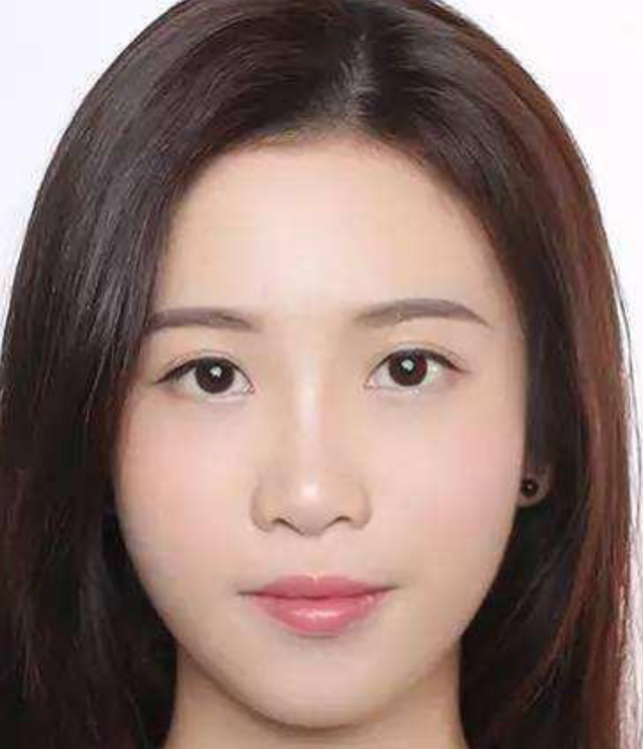}
        \caption{}
    \end{subfigure}
    \begin{subfigure}{0.2\linewidth}
        \includegraphics[width=\linewidth]{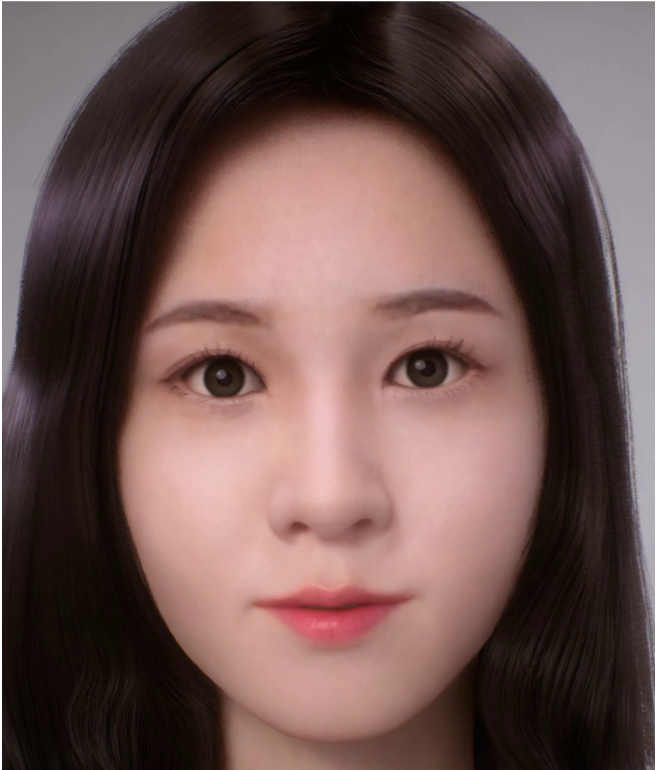}
        \caption{}
    \end{subfigure}
    \begin{subfigure}{0.2\linewidth}
        \includegraphics[width=\linewidth]{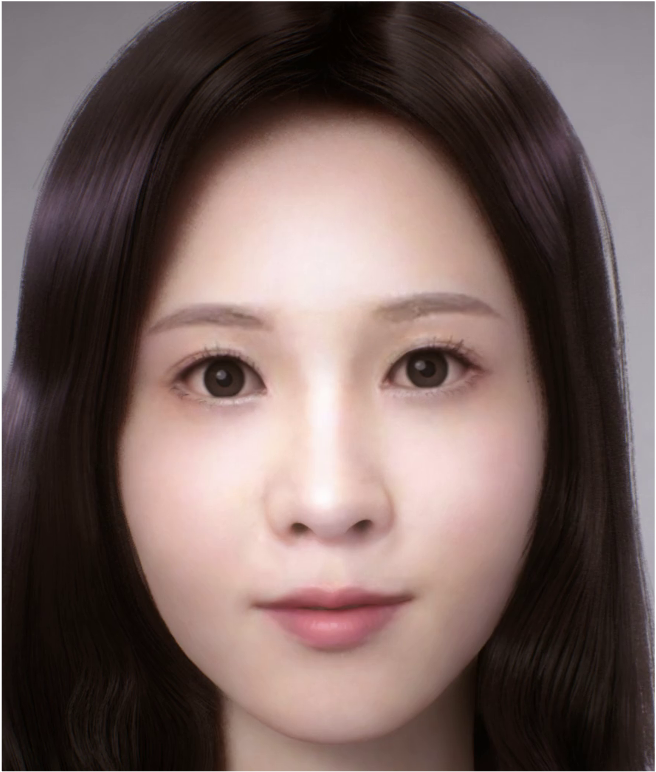}
        \caption{}
    \end{subfigure}
    \begin{subfigure}{0.2\linewidth}
        \includegraphics[width=\linewidth]{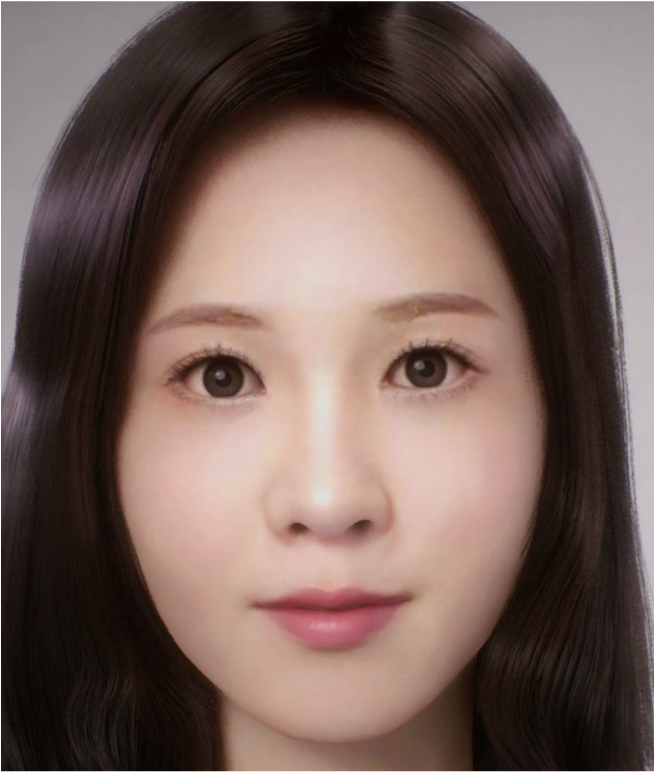}
        \caption{}
    \end{subfigure}
    \caption{Comparison of different types of textures rendered in UE. (a) is the input image, (b), (c), (d) are generated 3D characters using diffuse albedo~\cite{ren2023makeacharacter}, texture generated from differentiable rendering and the corrected textures.}
    \label{fig:color_correction_results}
    \vspace{-10pt}
\end{figure}

\subsection{Texture Generation}
\subsubsection{Texture Correction}

In our endeavor to accurately replicate the textures and shadings of the input facial images within Unreal Engine (UE), the illumination-independent diffuse albedos~\cite{ren2023makeacharacter}prove to be unsuitable for this scenario, as demonstrated in Figure~\ref{fig:color_correction_results}(b). While textures generated via differentiable rendering initially present as a promising choice, they inherently incorporate baked illumination, leading to significant discrepancies in light and shadow effects when rendered in UE compared to those observed in the input facial images, as depicted in Figure~\ref{fig:color_correction_results}(c). Moreover, recreating identical lighting conditions in UE is particularly challenging due to the inherently ill-posed problem of inferring illumination from a single real-world image. To address these issues, we opt to establish a fixed lighting environment within UE and formulate the task of lighting replication as a classical image-to-image translation problem. By learning the differences between the rendered output and the input facial image, it corrects the textures generated by the differentiable rendering and reproduces the lighting effects of the input images, as illustrated in Figure~\ref{fig:color_correction_results}(d).

\begin{figure}[tb]
\centering
\includegraphics[width=0.5\textwidth]{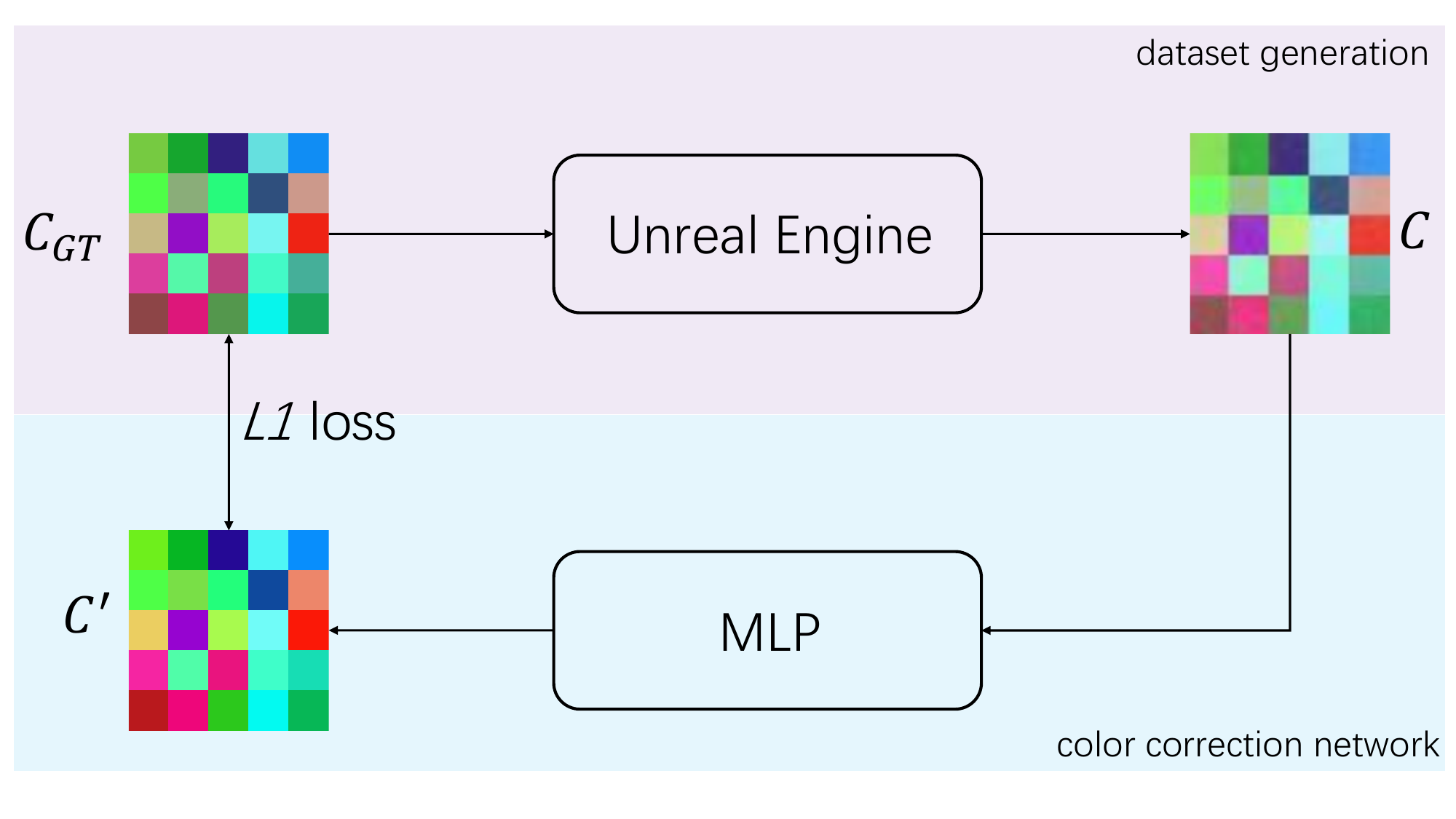}
\caption{The color correction pipeline.}
\label{fig:color_correction}
\end{figure}

It is well known that the rendering process in UE is not differentiable. To overcome this limitation and facilitate back-propagation of shadings onto facial textures, we propose a neural network, denoted as $\mathcal{N}$, to emulate the inverse rendering process of the UE. As in Figure~\ref{fig:color_correction}, $\mathcal{N}$ maps the rendered color $\mathcal{C}$ to a corrected color $\mathcal{C}^{'}$, such that
\begin{equation}
\mathcal{C}^{'}=\mathcal{N}(\mathcal{C})
\end{equation}
When the corrected color $\mathcal{C}^{'}$ is rendered through UE, the rendered result yields to $\mathcal{C}$, indicating successful replication of the rendered color. To achieve the replication of the input facial images' illumination, we apply $\mathcal{N}$ to adjust the textures generated by the differentiable rendering process.

We implement the network $\mathcal{N}$ as a three-layer MLP with 32 hidden units, and define the loss function as $\sum_{i=1}^{n}\|\mathcal{C}_{i}^{'}-\mathcal{N}(\mathcal{C}_{i})\|$. The training data is generated through an intuitive approach: we render a variety of colors $\mathcal{C}^{'}$ in UE and store the results as $\mathcal{C}$, yielding over 10,000 color pairs for training. 



\subsubsection{Facial Lighting Normalization}
\begin{figure}[tb]
    \centering
 
    \includegraphics[width=0.2\textwidth]{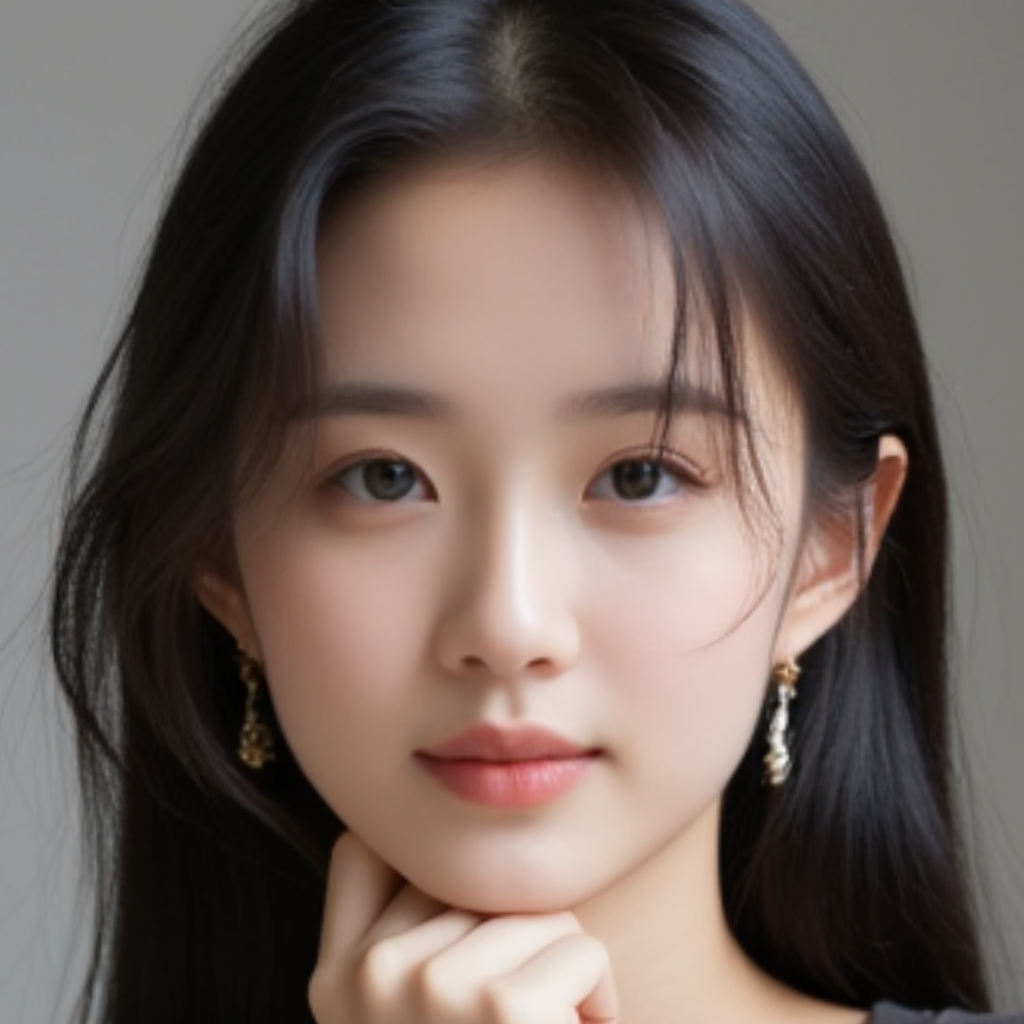}
    \includegraphics[width=0.2\textwidth]{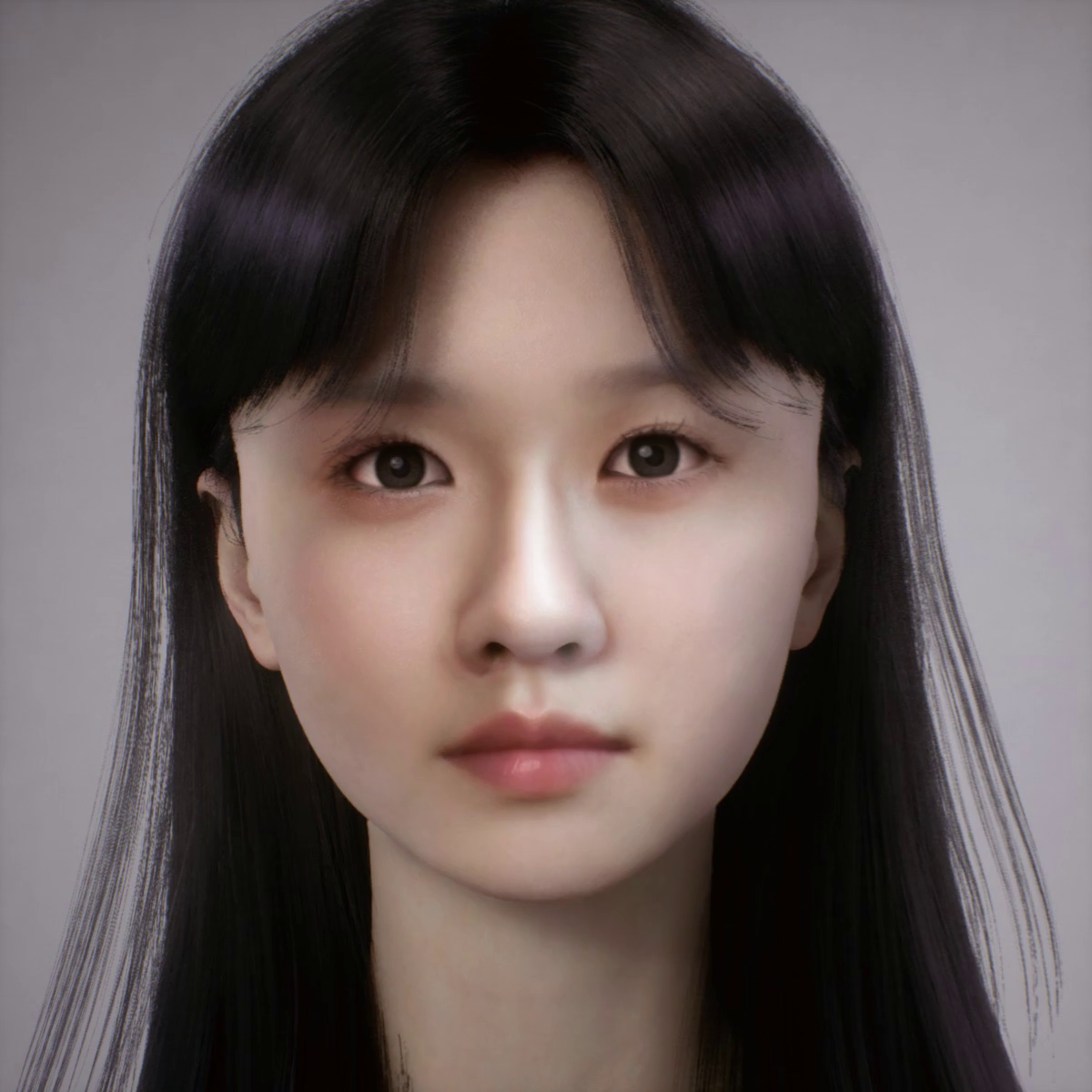}
    \includegraphics[width=0.2\textwidth]{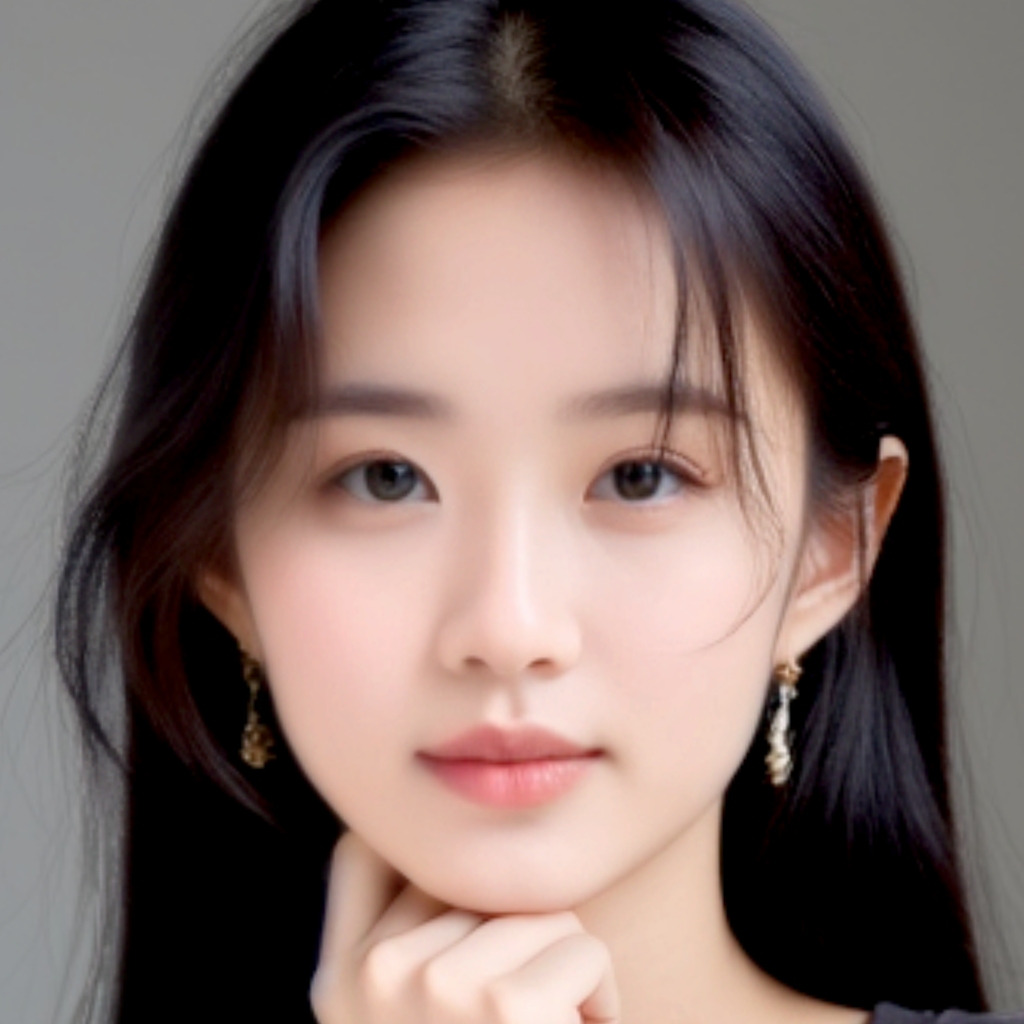}
    \includegraphics[width=0.2\textwidth]{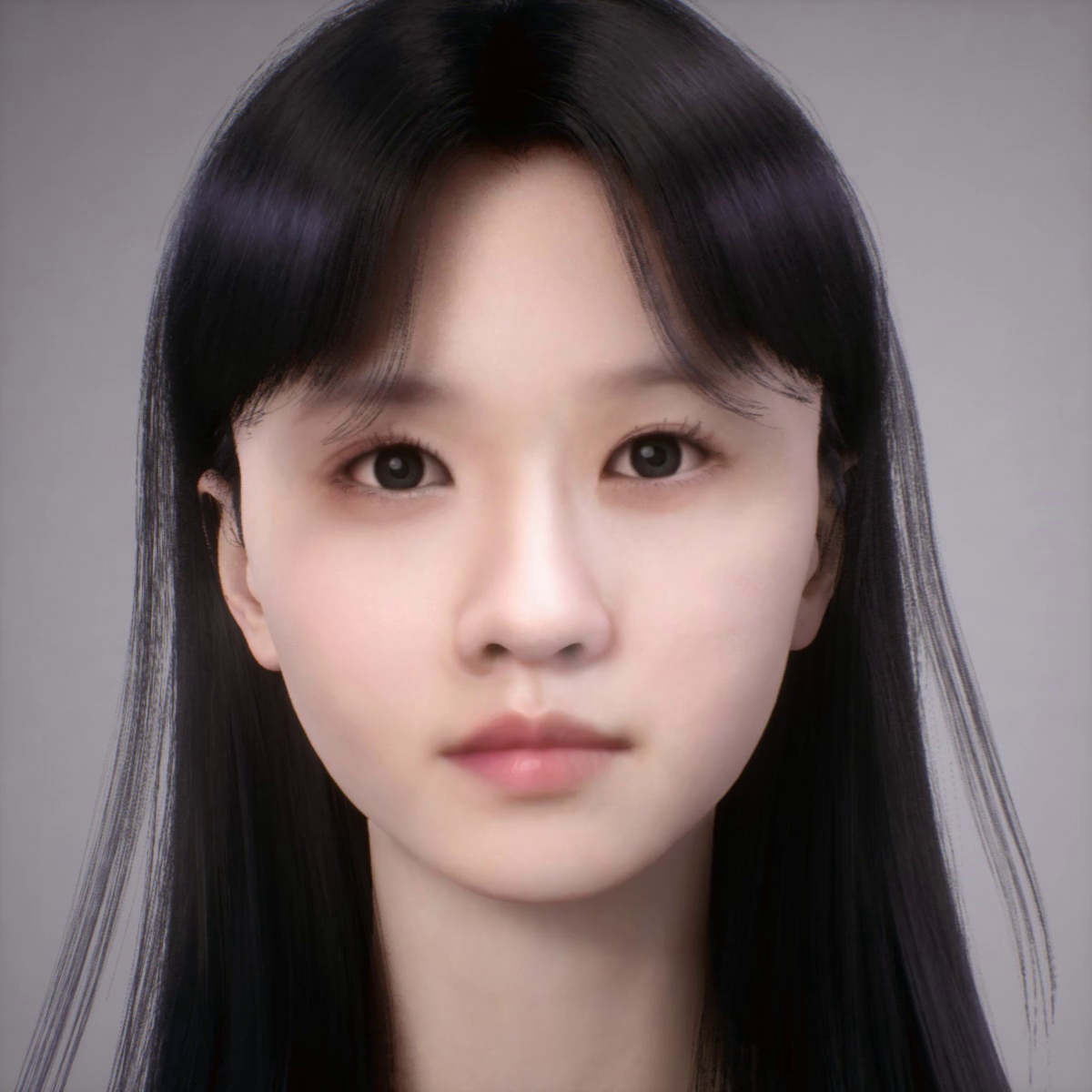}
  
    \caption{Results of the relighted image. This figure presents the 3D characters generated by the input facial image and the normalized facial image, respectively.}
\label{fig:relighted_results}
\end{figure}

When dealing with facial images under atypical lighting, such as non-uniform, colored, or overexposed lighting conditions, the corrected textures often retain these lighting features, including shadings and highlights. As a result, it presents a dilemma, while these lighting features enhance the fidelity of the rendered face compared to the input, they also create visual inconsistencies within the uniformly lit UE environment. To address this issue, we employ the IC-light algorithm \cite{iclight} to attenuate these atypical lighting features and achieve normalized facial lighting.
Given a facial image and the uniform light map, the IC-light algorithm re-lights the face according to the new lighting condition. To partially maintain the original shadings and highlights, we blend the re-lit face with the input face, yielding a similar yet even lighting facial image. Figure~\ref{fig:relighted_results}(a) and Figure~\ref{fig:relighted_results}(b) present the 3D characters generated by the input facial image  and the normalized facial image, demonstrating that the latter's shadings and highlights are more natural and harmonious within our UE environment.

\subsection{Hair Generation}



When generating 3D avatars from single images, capturing the hairstyle is crucial for achieving an accurate likeness. To facilitate this, we utilize a specialized hairstyle classification model powered by convolutional neural networks. This model directly maps a portrait image to a specific hairstyle within our asset library, bypassing the need for intermediate feature extraction. Many hairstyles naturally showcase asymmetry, as seen in the way long hair flows over the chest or back and the placement of bangs. In our asset library, we've enhanced these asymmetrical hairstyles by applying a horizontal flip, treating the flipped versions as unique styles. Consequently, when labeling the training data, we distinctly differentiate between the left and right variations of these hairstyles.

\section{3D Character Animation}
\label{sec:character_animation}
In our application we mainly concerned to animate the generated avatars via speech audio. We leveraged bone skeleton and facial control rig to generate gesture and facial motion, respectively.

\begin{figure*}[tb]
\centering
\includegraphics[width=0.85\textwidth]{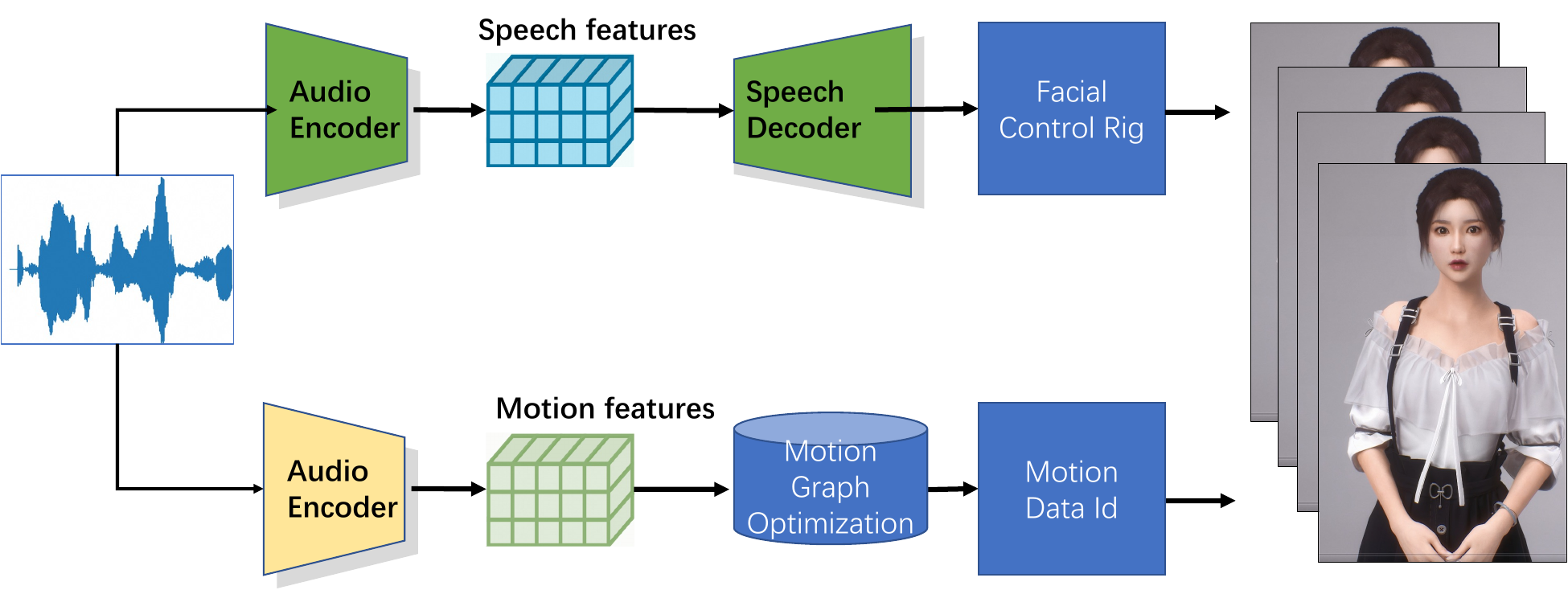}
\caption{Pipeline of animations. Coefficients of facial control rig were directly predicted from speech audio. Gesture motion clips were selected from the captured data set by optimizing graph path.}
\label{fig:animations}
\end{figure*}

\subsection{Speech-driven Gesture Motions}
Researchers have explored masked transformers ~\cite{Liu_2024_CVPR, Liu_2024_CVPRProbtalk, zhang2024semtalkholisticcospeechmotion} or diffusion models~\cite{ijcai2023p650, chen2024diffsheg, zhu2023taming} to predict gestures from speech with or without additional conditions, like word text, emotional labels or semantic labels. They all freely generate gestures by learning  distribution from motion capture data. Their jittering results do not meet the industrial requirements. In contrast, we suggest to predict predefined motion pieces for realistic animations.

In order to produce realistic animations, we use Vicon, which is a sophisticated motion capture system, to collect body motion data, and employ two senior animators to  adapt the data to our generated avatars. We capture more than one hundred pieces of motion data. The length of each piece of data falls between 2 seconds and 3 seconds. These data are common nonverbal gestures made by people while speaking, and they can be divided into 5 categories, which responds to 5 human poses(see Figure~\ref{fig:animations}). In each category, the start and end pose of each data are same. This allows the motion data in each category to be smoothly concatenated together to form longer animation. Besides, there is motion data for transition between different poses. These captured data form the foundation of our training data.
\begin{figure}[h]
\centering
\includegraphics[width=0.7\textwidth]{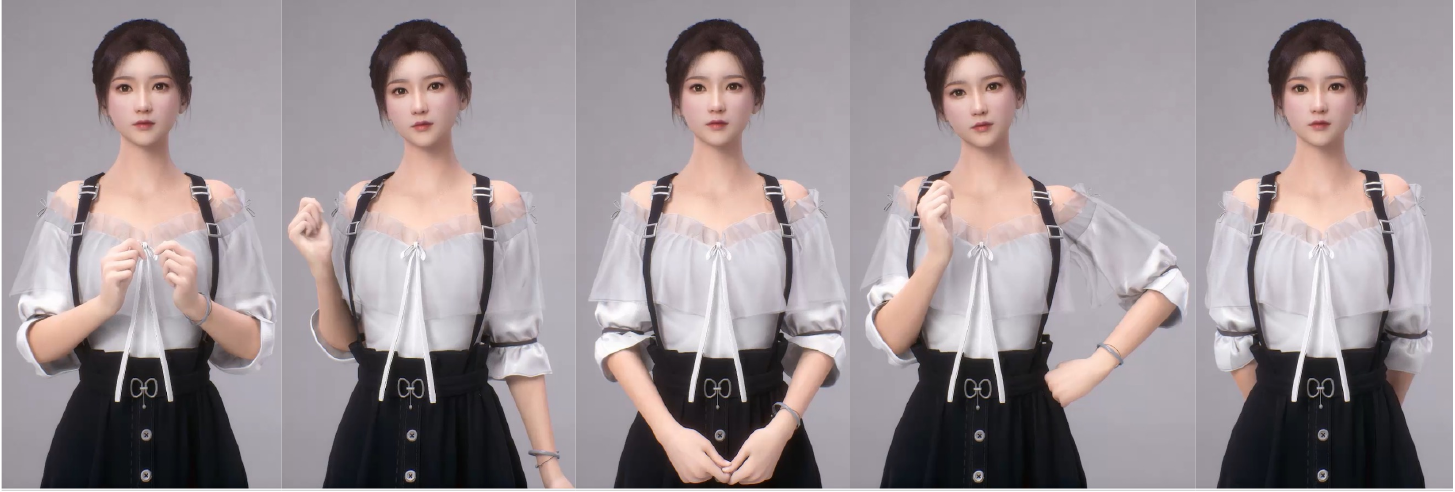}
\caption{Captured motion data can be divided into 5 categories corresponding to 5 poses. In each category, the start and end poses of each data are same. This enables us to generate longer animation by concatenating any two pieces of motion data in a single category.}
\label{fig:5poses}
\end{figure}

The animators generate sequences of motion data based on speech audios as our training data set. To align audio with motion sequences in the data set, we use graph-based optimization method to generate motion sequences. We take each piece of the data as a graph node, and there is a directed edge between two nodes if these two nodes are in continuous order in the training data set. The weight associated  with a edge is defined as follows:
\begin{equation}
T(N_i, N_j) = \lambda_1 T_p(N_i,N_j) + \lambda_2 T_r(N_i,N_j)
\end{equation}
where $N_i$ denotes the $i$th graph node, $T_p(N_i,N_j)$ is the translation loss between two nodes, $T_r(N_i,N_j)$ is the rotation loss of corresponding joints between two nodes. We also associate each node with an audio embedding feature which extracted from Wav2Vec2. Given an audio sequence $M_1, M_2,...., M_n$, the corresponding motion sequence is obtained by an optimal path in the directed graph. The cost is defined as follows:
\begin{equation}
C = \sum_i^n C_a(A_i, \hat{A}_i) + \sum_i^{n-1}T(i,i+1)
\end{equation}
where $C_a(A_i, \hat{A}_i)$ is the audio embedding loss. $A_i$ and $\hat{A}_i$ denote the  embedding  features of audio clip  $M_i$ and graph node $N_i$, respectively. $T(i,i+1)$ is the edge loss between the selected nodes with audio clip $M_i$ and $M_{i+1}$. We use The viterbi algorithm ~\cite{1450960} to solve this optimization problem. 

In application we optimize the graph path  within a specific motion data category. For lengthy speech audio, we randomly choose transition data to generate motions in another category. This strategy enables us to generate a wide range of realistic animations.

\subsection{Speech-Driven 3D Facial Animations}

\begin{figure}[tb]
\centering
\includegraphics[width=0.6\textwidth]{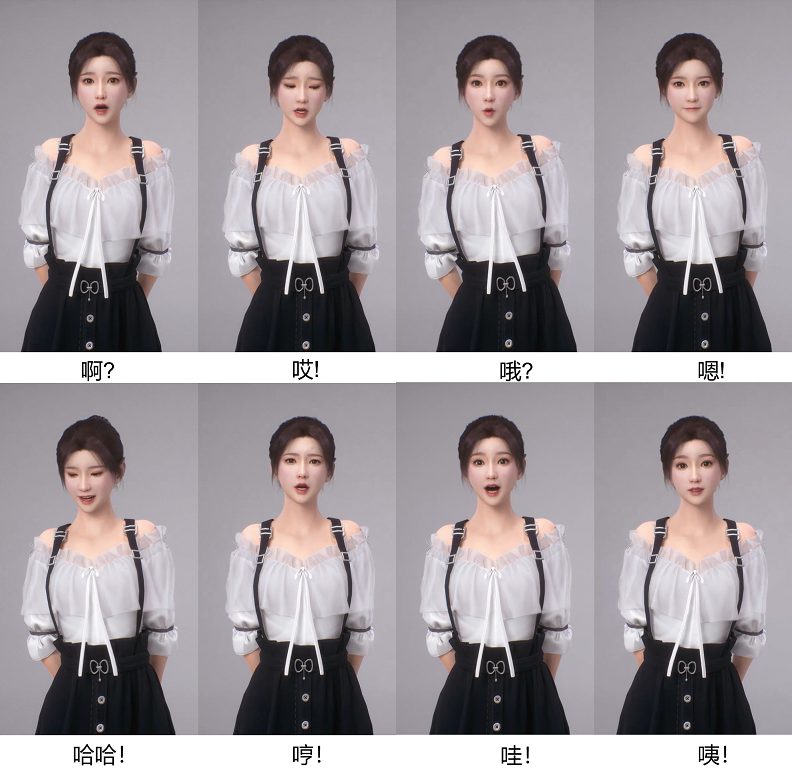}
  \vspace{-10pt}
\caption{Screenshots of some Chinese interjections template animations}
\label{fig:interjections}

\end{figure}

There are typically two parametric representation for creating 3D facial animations: vertex displacement and blendshapes. While vertex displacement methods can control facial movements with great precision, it is challenging to animate components such as teeth, tongue, and eyelashes, which can result in noticeable imperfections. On the other hand, blendshape-based solutions are stable and less prone to imperfections. However, the expressive capability of such solutions depends on the number of blendshapes available. A larger number of blendshapes allows for more nuanced expressions and improved animation quality, whereas a smaller number may result in less satisfactory outcomes. We adopt the blendshape-based approach, leveraging the robust Control Rig system in Metahuman \cite{metahuman}  to generate realistic facial animations. In order to obtain high-quality facial animations data, we employ a tool called MetaHuman Animator \cite{metahuman} to capture ctrlrig coefficients and then refine them with the expertise of experienced animators. 

The entire pipeline of speech-driven 3D Facial Animations is shown in Figure~\ref{fig:animations}. Firstly, we utilize the state-of-the-art self-supervised pre-trained speech model, wav2vec 2.0 \cite{baevski2020wav2vec}, to encode audio signals. Secondly, a non-autoregressive architecture is employed as the decoder to directly regress facial animation weights from audio features. In contrast to autoregressive transformer-based methods, the non-autoregressive method only call decoder once, so it more efficient than autoregressive method, which need a iterative decoding loop. We only employ reconstruction loss and velocity loss which is also used in Selftalk \cite{peng2023selftalk} to train the model. The loss function is formulated as:
\begin{equation}
L = L_{\text{rec}} + L_{\text{vel}}
\end{equation}

The reconstruction loss \( L_{\text{rec}} \) measures the difference between the predicted ctrlrig coefficients and the ground-truth ctrlrig coefficients. Here, \( b_{t,i} \) denotes  the ground truth coefficients of the  \( i^{\text{th}} \) ctrlrig at timestamp  \(t \), whereas \( \hat{b}_{t,i} \) represents the predicted value. \( T \) is the length of the sequence, \( N \) is the number of ctrlrig.
\begin{equation}
L_{\text{rec}} =   \sum_{t=1}^{T}  \sum_{i=1}^{N} \left\| b_{t,i} - \hat{b}_{t,i} \right\|_2^2
\end{equation}

The velocity loss  \( L_{\text{vel}} \) is used to reduce lip jittery over time.
\begin{equation}
L_{\text{vel}} = \sum_{t=1}^{T}  \sum_{i=1}^{N} \left\| (b_{t,i} - b_{t-1,i}) -  ( \hat{b}_{t,i} - \hat{b}_{t-1,i}) \right\|_2^2
\end{equation}

In order to enhance the facial expressions of digital avatars, we pre-create templates for words with strong emotions, such as interjections. ~\cref{fig:interjections} show some screenshots of the Chinese interjections we created. When these interjections are detected in the input audio, the corresponding facial expression templates are triggered to enhance the expression of emotions.

\section{Results}
\label{sec:Results}

\subsection{3D Character Generation}

\begin{figure*}[t]
  \centering
  \resizebox{0.9\linewidth}{!}{
   \includegraphics{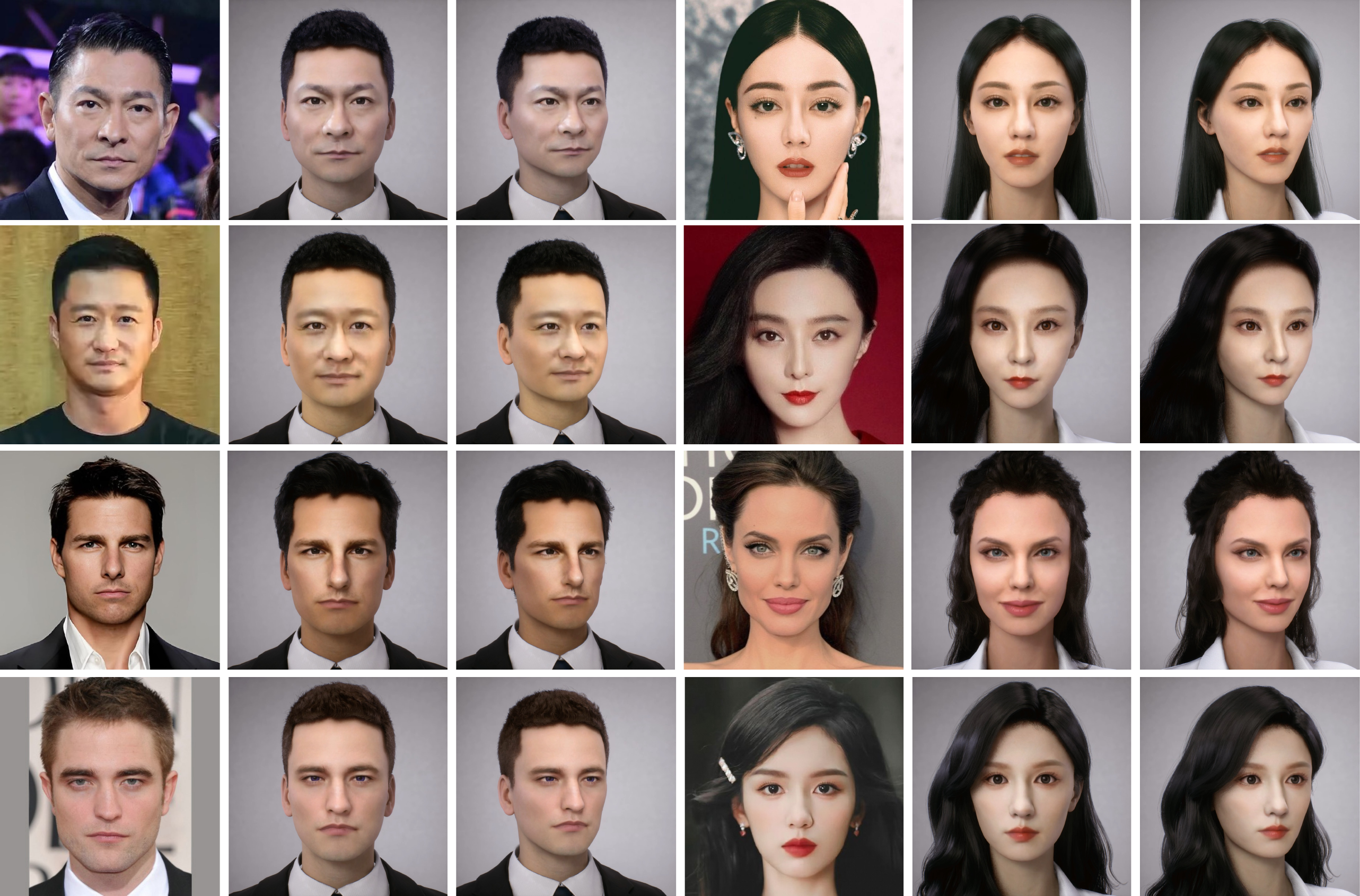} }
  \caption{Textured 3D character models generated from frontal portrait images. We utilize Unreal Engine for high-fidelity rendering.}
    
  \label{fig: showcases}
\end{figure*} 


Figure~\ref{fig: showcases} presents the 3D characters created from single images with our method. 
We compare our method with several state-of-art 3D avatar generation algorithms, including two text-based avatar generation methods(DreamFace~\cite{zhang2023dreamface}, Rodin~\cite{wang2022rodin}  ), and two single image-based avatar generation methods (MeInGame~\cite{Lin2022MeInGame} and MoSAR~\cite{Dib2023MoSAR}).
As shown in Table~\ref{tab:compared}, Our method generates complete, production-ready 3D character assets compatible with modern CG pipelines, offering enhanced rendering fidelity, highly accurate rigging, and improved editability compared to other approaches.

\begin{table*}[h!]

  \centering
  \begin{minipage}{0.8\linewidth}
    \begin{tabular}{@{}lcccccc@{}} 
    \hline
    \textbf{Method} & \textbf{High Fidelity} & \textbf{Full-Body}  & \textbf{Facial Rigging} & \textbf{CG Engine} & \textbf{Outfit}\\
     & \textbf{Generation} & \textbf{Completeness} & \textbf{Fidelity}  & \textbf{Compatible} & \textbf{Editable}\\
    \hline
    DreamFace~\cite{zhang2023dreamface}  & $\star$ $\star$ $\star$ & $\times$ & $\star$ & $\checkmark$ & $\checkmark$ & \\
    Rodin~\cite{wang2022rodin}  & $\star$ $\star$ & $\times$ & $\times$ & $\times$ &  $\times$ & \\
    MeInGame~\cite{Lin2022MeInGame} &$\star$ & $\checkmark$ & $\star$ & $\checkmark$ & $\checkmark$ & \\
    MoSAR~\cite{Dib2023MoSAR} & $\star$ $\star$ & $\times$ & $\star$ & $\checkmark$ & $\checkmark$ & \\
    Ours & $\star$ $\star$ $\star$ & $\checkmark$ & $\star$ $\star$ $\star$ & $\checkmark$ & $\checkmark$ &  \\
    \hline
    
    \end{tabular}
    \caption{Comparison of latest related 3D avatar generation algorithms.}
       \label{tab:compared}
    \end{minipage}
    
\end{table*}

\begin{figure*}[h]
\centering
\includegraphics[width=0.98\textwidth]{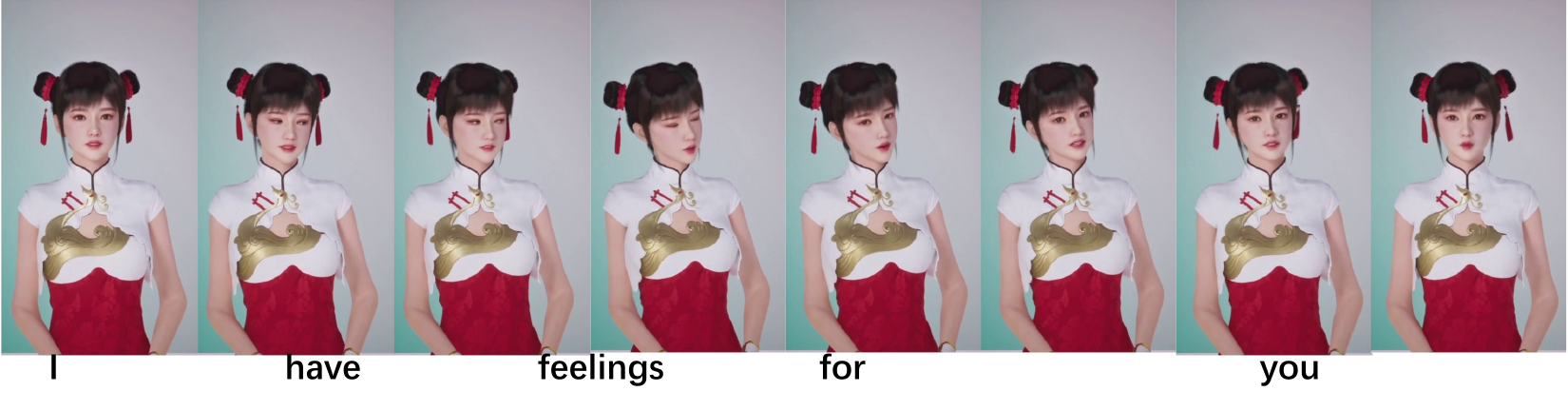}
\includegraphics[width=0.98\textwidth]{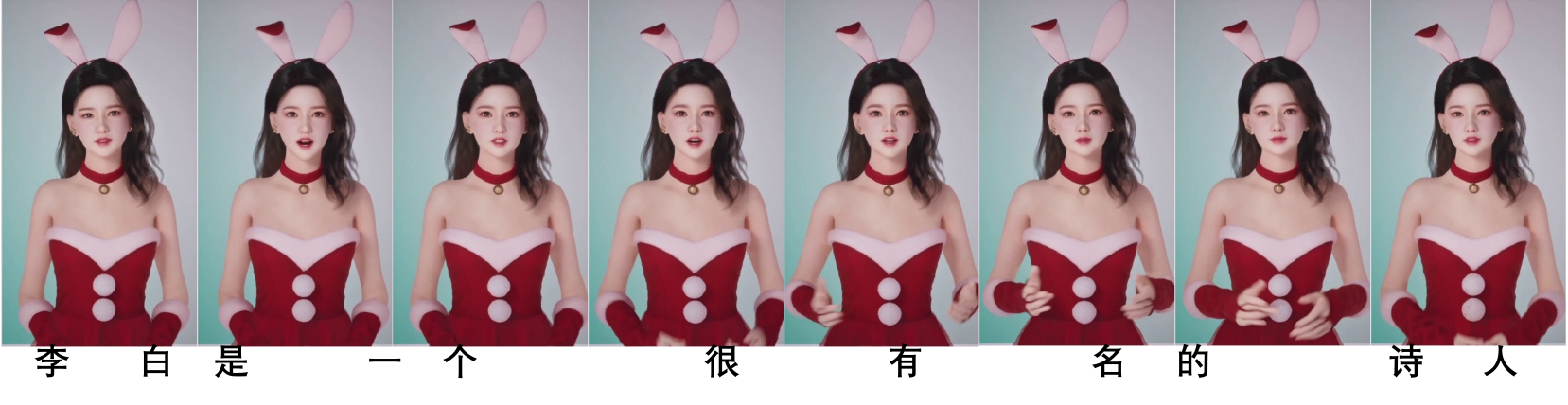}
\includegraphics[width=0.98\textwidth]{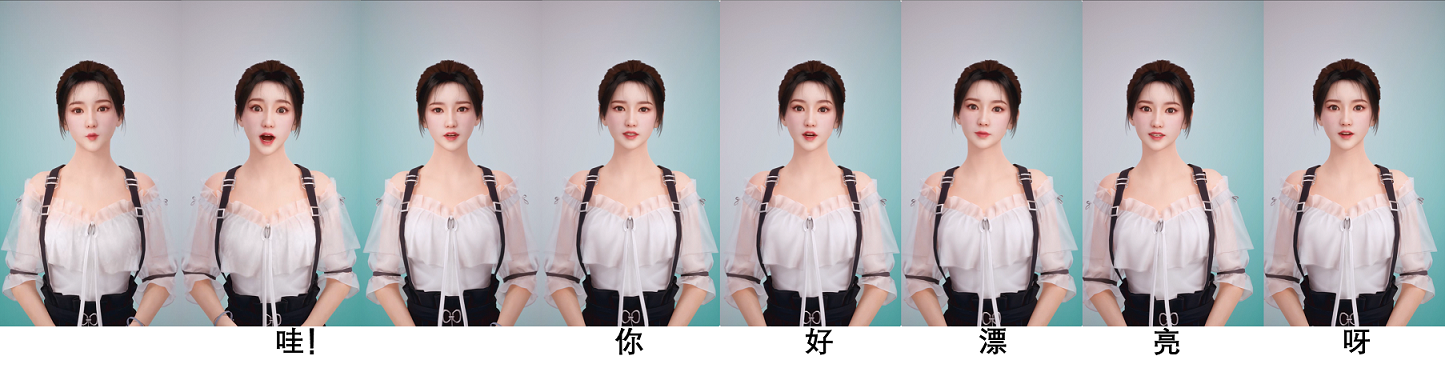}
  \vspace{-10pt}
\caption{Full-body animations sequences generated by our system with english speech(top row) ,Chinese speech(middle row) and Chinese speech with interjection(bottom row).}
\label{fig:libai}
\end{figure*}

\subsection{3D Character Animation}
Our system combines the facial animation and gesture animation, it can generate holistic full-body motions. Figure~\ref{fig:libai} shows
holistic gesture motions and facial animations generated from English and Chinese speeches. Benefit from high-quality mouth animations data, and non-autoregressive 
 decoder architecture, our system can generate realistic lip-sync with efficiency exceeding real-time performance. Additionally, suitable expressions are also incorporated into facial animations during speech to enhance emotions. Our predefined motion data set guarantees no jittering gestures. And the gesture motions are aligned well with speech rhythm(See results  in appendix videos).

\section{Limitation}
\label{sec:Limitation}
Although our method excels in generating high-quality 3D character assets and animations, it still has several limitations. Firstly, when the input face does not have a neutral expression, our expression correction process may introduce artifacts, leading to unsatisfactory facial rig results. Additionally, our generated characters currently lack the ability to perform complex dynamic animations, such as singing or dancing, due to a lack of specific training data. Furthermore, the general nature of our animation model limits the personalization of facial and body motion synthesis. We are actively researching ways to overcome these limitations.

{\small
\bibliographystyle{ieeenat_fullname}
\bibliography{_main}
}


\end{document}


\title{\paperTitle}
\author{\authorBlock}
\maketitlesupplementary

\appendix
\section{Appendix Section}
\label{sec:appendix_section}
Supplementary material goes here.

{\small
\bibliographystyle{ieeenat_fullname}
\bibliography{11_references}
}